\documentclass[runningheads]{llncs}
\usepackage[T1]{fontenc}
\usepackage{graphicx}
\usepackage{booktabs}
\begin{document}

\raggedbottom
\title{Brain-Aligned Multimodal LLMs for Social and Emotional Understanding of Cartoon Movies}

% \title{Can Multimodal LLMs Track Theory of Mind and Valence?}

%
\titlerunning{Brain-Aligned MLLMs for Social-Emotional Movie Understanding}
\author{Yiran Huang\inst{1}\orcidID{0009-0005-6005-0186} \and
Zeyong Zhang\inst{1}\orcidID{0009-0004-5855-0520} \and
Bharat Biswal\inst{2}\orcidID{0000-0002-3710-3500}\and
Xin Di\inst{2}\orcidID{0000-0002-2422-9016} \and
Mengjia Xu\inst{1}$^{*}$\orcidID{0000-0002-3627-8261}}
%0
\authorrunning{Yiran H. et al.}

\institute{$^{*}$Department of Data Science, New Jersey Institute of Technology, Newark, NJ 07102, USA \\
\email{yh87@njit.edu, zz63@njit.edu, mx6@njit.edu} \and
Department of Biomedical Engineering, New Jersey Institute of Technology, Newark, NJ 07102, USA \\
\email{bharat.biswal@njit.edu, xin.di@njit.edu}
}

\begingroup
\renewcommand\thefootnote{}
\footnotetext{$^{*}$Corresponding author}
\endgroup
\maketitle              % typeset the header of the contribution
\begin{abstract}
Multimodal large language models (MLLMs) offer scalable annotations of social and affective content, but their temporal agreement with human judgments and neural predictive value require further evaluation. We compared diffusion-based and autoregressive MLLMs with continuous human Theory of Mind (ToM) and affective valence ratings across three short cartoon movies. Behavioral analyses assessed temporally aligned Pearson correlations, rating distributions, and the effects of preceding scene descriptions. Ratings from each source were separately used in subject-specific regression models to predict regional fMRI responses.
Model ratings captured aspects of human temporal variation despite systematic distributional differences, with generally stronger alignment for valence than ToM. Preceding context improved alignment in some conditions, depending on the model and movie. In \emph{The Present}, more regions showed significant encoding for MLLM ratings than for human ratings, whereas effects were more limited in the other movies. Spatial overlap across rating sources occurred mainly among MLLMs; overlap between human and model ratings was limited and involved primarily visual, dorsal-attention, and posterior default-mode regions.
These findings support MLLMs as scalable tools for generating brain-predictive annotations of social and emotional content, while highlighting systematic rating differences and model- and stimulus-dependent alignment with human judgments.

% (https://github.com/Yiran-H/functional-MRI-Movie-MLLMs)

\keywords{Functional MRI  \and Theory of Mind \and Affective Valence \and Multimodal Large Language Models \and Naturalistic Neuroimaging.}
\end{abstract}

\section{Introduction}

Understanding how the brain processes unfolding social and emotional information remains a fundamental challenge in cognitive neuroscience. Naturalistic movies provide a framework for studying these processes through rich visual, auditory, and narrative content within a shared temporal structure. Movie viewing elicits reliable, stimulus-locked neural responses across individuals, including in higher-order association cortex \cite{hasson2004intersubject,hasson2010reliability}. It also evokes intersubject-consistent patterns of dynamic functional connectivity that differ across movies, suggesting that movie content engages distinct patterns of large-scale neural coordination \cite{di2022dynamic}. In developmental neuroimaging, engaging movies facilitate data collection while reducing explicit task demands \cite{vanderwal2019movies}. Together, these properties make movie viewing a promising paradigm for examining how time-varying social and emotional information relates to brain activity.

% MX: add 1-2 references in this paragraph 

Theory of Mind (ToM), reasoning about others' beliefs, intentions, and desires \cite{saxe2013people}, and valence, the positive--negative dimension of affective experience \cite{russell1980circumplex}, are especially relevant to naturalistic social cognition.
Both evolve continuously as characters interact and a scene's social and emotional meaning unfolds, so naturalistic paradigms offer a way to move beyond discrete task events and treat ToM and valence as continuous, time-varying features of an ongoing stimulus.
Movie paradigms have been widely used to investigate social and emotional processing. For example, viewing of the animated movie \emph{Partly Cloudy} 
engages the developing ToM network in children and elicits temporally predictive responses to mental-state information \cite{richardson2018development,richardson2020development}, and continuous emotional annotations have been used to characterize the neural representation of emotion during movie viewing \cite{saarimaki2025cerebral}.  A central challenge, however, is obtaining annotations that capture these continuously varying psychological dimensions: conventional ToM assessment relies on heterogeneous tasks with substantial psychometric variability \cite{beaudoin2020systematic}, and continuous emotion rating demands sustained attention and effort. Naturalistic neuroimaging studies therefore typically collect time-varying annotations from independent raters as shared stimulus features, preserving the low task demands of passive viewing \cite{vanderwal2019movies}, but such annotation remains labor-intensive, motivating more scalable alternatives.

% Background of MLLMs in Neuroscience
Multimodal large language models (MLLMs) have recently emerged as a complementary approach that addresses these limitations by combining visual perception, instruction following, and semantic reasoning within a unified architecture \cite{caffagni2024revolution}. The MLLM-generated emotional ratings of static images correlate with normative human ratings~\cite{romeo2025artificial}, yet their relationship to brain responses remains unexplored. Video-based ToM benchmarks have also evaluated MLLMs' ability to infer characters' mental states through question answering, rather than continuous ratings \cite{villa2025moments}. While such benchmarks provide a structured evaluation of ToM reasoning, they do not capture the continuous temporal dynamics of social and emotional content. 
It therefore remains unclear how closely MLLM-generated ratings track human continuous judgments during movie viewing, and how temporal context influences this alignment. 
Moreover, prior works largely focused on autoregressive MLLMs (e.g., GPT-4o, Gemini Pro, Claude Sonnet), with little attention to diffusion-based MLLMs (e.g., LLaDA-V~\cite{you2026llada}).  

% From continuous ratings to brain encoding
Beyond behavioral agreement, MLLM-generated annotations can also be evaluated by their ability to predict brain activity. Previous studies have used human-annotated social and emotional features to predict movie-evoked BOLD responses \cite{masson2021functional,im2025early}; complementary approaches have modeled temporal dependencies in functional connectivity to characterize evolving brain interactions \cite{huang2025brainatcl}. Visual and language-model representations have also shown predictive value for naturalistic neural responses \cite{small2024vision,nakagi2024unveiling}. 
% However, the relationship between MLLM-generated ratings and movie-evoked fMRI responses remains less explored.
% More recently, psychologically interpretable movie annotations generated by Gemini were shown to predict responses across distributed association cortex \cite{li2026mapping}, using Gemini 2.5 Pro~\cite{comanici2025gemini} on audiovisual clips to generate brain-predictive annotations, while the model is not open-weight and human ratings cover only part of the movie. 
However, how MLLM-generated social and emotional ratings relate to neural activity remains underexplored. Recent work showed that interpretable annotations generated by Gemini 2.5 Pro \cite{comanici2025gemini} from audiovisual movie clips predicted responses across distributed association cortex \cite{li2026mapping}, but relied on a proprietary model, limiting reproducibility, and validated the annotations against human ratings for only part of the movie.

% We focus on all open-weight MLLMs spanning autoregressive and diffusion-based architectures, enabling reproducible and controlled evaluation of model-generated annotations. Specifically, we consider Qwen3-VL~\cite{bai2025qwen3} and Gemma 3~\cite{team2025gemma} as autoregressive MLLMs and LLaDA-V as a diffusion-based MLLM\cite{you2026llada}.

\begin{figure}[t]
\includegraphics[width=\textwidth]{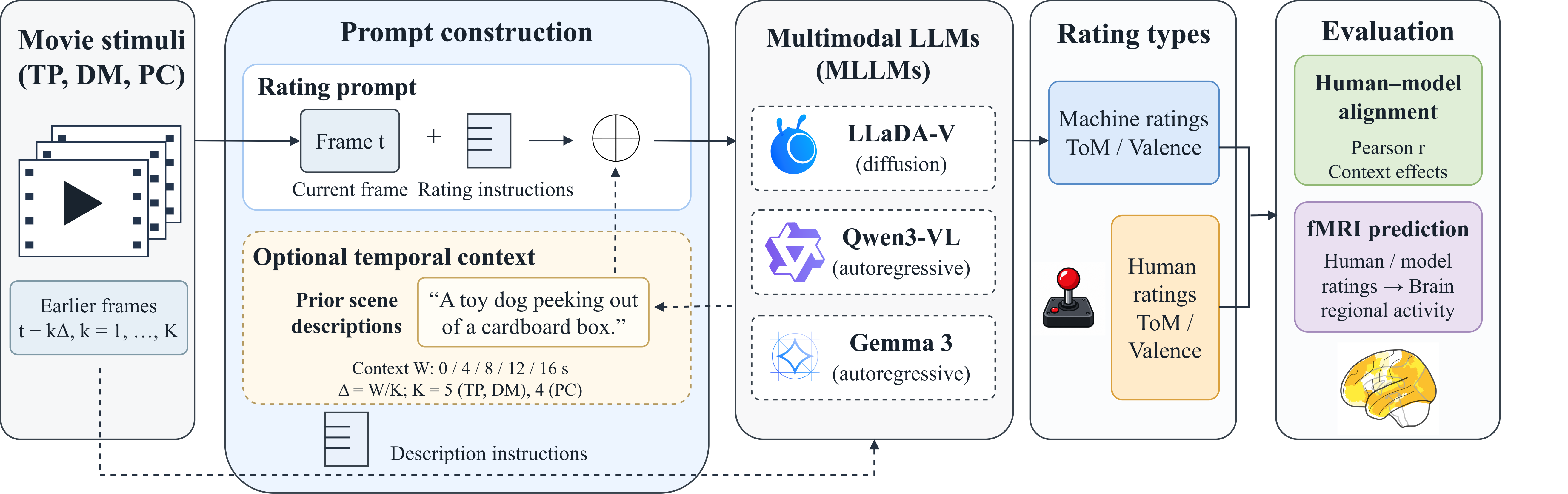}
% \caption{
% Overview of the brain-alignment framework for evaluating multimodal LLMs (MLLMs) on social and emotional understanding of cartoon movies. Detailed prompt instructions can be found from our project github repo.  \url{https://github.com/Yiran-H/functional-MRI-Movie-MLLMs}
% % Movie stimuli are processed by multiple MLLMs under varying temporal-context conditions to generate ToM and valence ratings, which are compared with human annotations and used to predict fMRI responses.
% }
\caption{Overview of the brain-alignment framework for evaluating multimodal LLMs (MLLMs) on social and emotional understanding of cartoon movies. Detailed prompts are available at \url{https://github.com/Yiran-H/functional-MRI-Movie-MLLMs}}
\label{fig0}
\vspace{-.19in}
\end{figure}

To address these gaps, our work  (Fig. \ref{fig0}) compares human ratings with continuous, frame-wise ToM and valence ratings generated by open-weight autoregressive and diffusion-based MLLMs (i.e., Qwen3-VL, Gemma 3, and LLaDA-V) across three animated movies. 
% \emph{The Present}, \emph{Despicable Me}, and \emph{Partly Cloudy}. 
We first evaluate human--model agreement, rating distributions, inter-model agreement, and the effect of preceding semantic scene descriptions on rating alignment. We then use ratings from each source as predictors in cross-validated fMRI encoding models to identify brain regions with significant predictive performance and assess their spatial overlap. Together, these analyses characterize the behavioral and neural alignment of MLLM-generated social and emotional annotations during naturalistic movie viewing, providing a framework for evaluating open-weight MLLMs as scalable, reproducible annotators of dynamic human-centered content.

\vspace{-.17in}

\section{Methods}

\vspace{-.03in}
\subsection{Datasets and Participants}

% We analyzed fMRI data from three naturalistic movie-watching datasets: \textit{The Present} (TP), \textit{Despicable Me} (DM), and \textit{Partly Cloudy} (PC). TP (3.3 min) and DM (10 min) were obtained from the Healthy Brain Network~\cite{alexander2017open}.
% After strict quality control, 104 unique participants (61 males and 43 females; aged 5.0 -- 21.9 years, mean $\pm$ SD = 12.0 $\pm$ 4.1 years) were included in the final analysis. Moreover, the PC dataset was obtained from the OpenNeuro database~\cite{ds002886:1.1.0}. Participants watched a 5.6-minute silent version of PC. Following quality control, the final sample consisted of 82 participants.

We analyzed fMRI data from three naturalistic movie-watching datasets: \textit{The Present} (TP), \textit{Despicable Me} (DM), and \textit{Partly Cloudy} (PC). TP (3.3 min) and DM (10 min) were obtained from the Healthy Brain Network~\cite{alexander2017open}; after quality control, 87 participants from the TP dataset and 83 from the DM dataset remained. In total, 104 unique participants were included in the final analysis, with 66 individuals appearing in both datasets. The final cohort comprised 61 males and 43 females, aged 5.0--21.9 years (mean $\pm$ SD = 12.0 $\pm$ 4.1 years).
PC was obtained from OpenNeuro~\cite{ds002886:1.1.0}: participants watched a 5.6-minute silent version.
Following quality control, the final sample consisted of 82 participants (45 females and 37 males; aged 3.5--39.0 years, mean $\pm$ SD = 13.2 $\pm$ 9.2 years), comprising an adult group of 29 participants (17 females and 12 males; aged 18--39 years, mean $\pm$ SD = 24.6 $\pm$ 5.3 years) and a child group of 53 participants (28 females and 25 males; aged 3.5--12.3 years, mean $\pm$ SD = 7.0 $\pm$ 2.5 years).

\vspace{-.15in}

\subsubsection{fMRI Data Preprocessing and Time-Series Extraction}
% We used residual fMRI time series preprocessed as described previously for PC~\cite{di2022principal} and TP/DM~\cite{di2023individual}. Functional images were realigned, coregistered to individual T1-weighted images, normalized to Montreal Neurological Institute (MNI) space at 2.4-mm isotropic resolution for TP/DM and 3-mm for PC, and smoothed with an 8-mm Gaussian kernel. Participants exceeding head-motion thresholds of 2.4 mm/2.4$^\circ$ for TP/DM or 1.5 mm/1.5$^\circ$ for PC were excluded. A voxel-wise general linear model~\cite{penny2011statistical} removed nuisance effects using Friston's 24 motion parameters~\cite{friston1996movement} and a 1/128-Hz high-pass filter to remove low-frequency drifts. The residual time series were temporally aligned with the movie features. After discarding the first 10 time points, PC retained 168 samples (TR $=2.0$~s), and TP and DM retained 240 and 740, respectively (TR $=0.8$~s). Voxel-wise responses were averaged within 100 Schaefer cortical parcels~\cite{schaefer2018local} and 14 AAL subcortical ROIs~\cite{tzourio2002automated}.

We used residual fMRI time series preprocessed as described previously for PC~\cite{di2022principal} and TP/DM~\cite{di2023individual}.
% Functional images were realigned, coregistered, normalized to MNI space, and smoothed with an 8-mm Gaussian kernel. Participants exceeding dataset-specific motion thresholds were excluded. Nuisance effects were removed using Friston's 24 motion parameters~\cite{friston1996movement} and a 1/128-Hz high-pass filter. After discarding the first 10 time points, PC retained 168 samples (TR $=2.0$ s), while TP and DM retained 240 and 740 samples, respectively (TR $=0.8$ s). 
The analyses included 114 ROIs: 100 cortical regions from the Schaefer atlas~\cite{schaefer2018local} and 14 subcortical regions from the AAL atlas~\cite{tzourio2002automated}.

\vspace{-.17in}

% \subsubsection{Continuous Subjective Ratings}

% Continuous subjective ratings were collected from two independent cohorts of healthy participants at the New Jersey Institute of Technology (NJIT) to characterize moment-to-moment fluctuations in perceived movie content. The Theory of Mind (ToM) cohort included 34 participants (17 females; age range: 18--27 years; mean $\pm$ SD: 20.9 $\pm$ 2.3 years), and the affective-valence cohort included 42 participants (8 females; age range: 17--28 years; mean $\pm$ SD: 19.8 $\pm$ 2.3 years). ToM data and procedures were reported previously~\cite{di2026beyond}. All participants provided informed consent under an IRB-approved protocol, reported normal or corrected-to-normal sensory function, and had no history of neurological or psychiatric disorders.

% After brief practice, participants watched the same three fMRI clips in a controlled laboratory setting. Continuous ratings were recorded at 4 Hz using CARMA software~\cite{girard2014carma} and a gaming joystick, with its vertical position mapped to a scale from $-10$ to $10$. 
% The ToM cohort continuously rated the extent to which they perceived or inferred characters' mental states (e.g., emotions, thoughts, and intentions), from ``not at all'' ($-10$) to ``very strongly'' ($10$). Using the same setup, the affective-valence cohort rated each clip's moment-to-moment valence, from highly negative ($-10$) to highly positive ($10$).

\subsubsection{Continuous Subjective Ratings}

Continuous subjective ratings were collected from two independent cohorts of healthy NJIT participants. The ToM cohort included 34 participants (17 females; age $20.9\pm2.3$ years), and the valence cohort included 42 participants (8 females; age $19.8\pm2.3$ years). ToM data and procedures were reported previously~\cite{di2026beyond}. All participants provided informed consent under an IRB-approved protocol and reported no neurological or psychiatric disorders.
Following brief practice, participants watched the same three movie clips in a controlled laboratory setting. Ratings were recorded at 4 Hz using CARMA software~\cite{girard2014carma} and a gaming joystick, with responses mapped to $[-10,10]$. The ToM cohort rated perceived characters' mental states, from ``not at all'' ($-10$) to ``very strongly'' ($10$), while the valence cohort rated moment-to-moment affective valence, from highly negative ($-10$) to highly positive ($10$).

\vspace{-.17in}

% MX revised this section! 9/16/26
\subsection{Automated ToM and Valence Rating with Multimodal LLMs}
We focused on three open-weight MLLMs spanning autoregressive and diffusion-based architectures to generate ToM and valence ratings from three aforementioned movies, enabling reproducible and controlled evaluation of model-generated annotations. Specifically, we used Qwen3-VL~\cite{bai2025qwen3} and Gemma 3~\cite{team2025gemma} as autoregressive MLLMs and LLaDA-V~\cite{you2026llada} as a diffusion-based MLLM. These models span different model families and generation paradigms: the diffusion-based LLaDA-V (8B) generates text via iterative denoising of masked tokens, while Qwen3-VL-8B-Instruct and Gemma 3 12B-IT are autoregressive, instruction-tuned models that generate tokens sequentially. We selected the Gemma 3 12B variant as the closest available model size to the other models.
% All models were decoded deterministically: Qwen3-VL and Gemma 3 used greedy decoding, and LLaDA-V used its default low-confidence remasking with temperature 0, 128 denoising steps, and a generation length of 128 tokens.

\vspace{-.2in}

\subsubsection{No-Context Baseline} 
In the baseline, each sampled movie frame was presented independently. Frames were sampled every 0.8 s for TP and DM and every 1.0 s for PC. All models received the same prompt template and rating instructions, requesting ToM and valence scores for the visible character(s) on a scale from $-10$ to $+10$, consistent with the human rating scales. No preceding frames or narrative information were provided. Ratings were generated frame by frame for each movie, smoothed with a Gaussian kernel ($\sigma = 2$ s), and subsequently aligned to its fMRI repetition time (TR) grid for comparison with human ratings. The baseline ratings were used for all fMRI encoding analyses; the context conditions described below were used only to assess how context affected agreement with human ratings.

% \vspace{-.2in}

\subsubsection{Temporal Context Integration}

To test the effect of preceding narrative context, the MLLMs first generated scene descriptions of frames sampled every 0.8~s for TP and DM and 1~s for PC. Descriptions of preceding frames were then added to the rating prompt alongside the current frame. We held the number of descriptions fixed (five for TP and DM; four for PC) and spaced them evenly across lookback windows of 4, 8, 12, and 16~s.
\vspace{-.2in}

\subsubsection{Comparison with Human Ratings}

Human ToM and valence ratings were screened for each movie and rating measure using a leave-one-out reliability procedure: each rater's time series was correlated with the mean of the remaining raters, and the $20\%$ raters with the lowest correlations were excluded. This retained 28 of 34 ToM raters and 34 of 42 valence raters for each movie. Ratings from the remaining raters were averaged for each movie and rating measure, then resampled to the fMRI TR grid using linear interpolation without additional smoothing.

\begin{table}[t]
\centering
\caption{Human--model Pearson correlations without temporal shift ($r_0$) and at the lag maximizing correlation ($r^{*}$, exploratory). Lags (in seconds, in parentheses) indicate model ratings leading human ratings.}
\label{tab:lags}
\begin{tabular}{|l|l|c|c|c|c|c|c|}
\hline
 & & \multicolumn{2}{c|}{LLaDA-V} & \multicolumn{2}{c|}{Qwen3-VL} & \multicolumn{2}{c|}{Gemma 3} \\
\cline{3-8}
Measure & Movie & $r_0$ & $r^{*}$ (lag) & $r_0$ & $r^{*}$ (lag) & $r_0$ & $r^{*}$ (lag) \\
\hline
ToM     & TP & .418 & .418 (0.0) & .063 & .063 (0.0) & .292 & .331 (1.6) \\
                         & DM & .034 & .054 (2.4) & .213 & .252 (3.2) & $-$.078 & $-$.051 (6.4) \\
                         & PC & .366 & .428 (4.0) & .523 & .609 (4.0) & .240 & .277 (4.0) \\
\hline
Valence & TP & .548 & .554 (0.8) & .401 & .411 (0.8) & .370 & .393 (1.6) \\
                         & DM & .416 & .423 (4.0) & .641 & .662 (4.8) & .615 & .639 (4.8) \\
                         & PC & .185 & .256 (4.0) & .060 & .141 (4.0) & .408 & .424 (2.0) \\
\hline
\end{tabular}
\end{table}

Model ratings, which fluctuated frame to frame, were Gaussian-smoothed ($\sigma = 2.0$~s) and resampled onto the human-rating TR grid.
Continuous joystick ratings are expected to lag behind the perceived content, whereas models rate each presented frame directly. To account for this human response delay, we compared model ratings at time $t$ with human ratings at time $t + L$, i.e., allowing model ratings to lead human ratings by a lag $L$. Lags were searched in one-TR steps from 0 to 6.4~s for TP and DM (TR $= 0.8$~s) and from 0 to 6.0~s for PC (TR $= 2.0$~s). For each movie, model, rating measure, and context condition, we selected the lag that maximized the Pearson correlation with human ratings. Because lag selection and evaluation used the same data, these optimized correlations are upwardly biased and are reported as exploratory, alongside the unshifted ($L = 0$) correlations. For the baseline condition, lag optimization increased correlations by at most 0.09 (median 0.025) relative to the unshifted values (Table~\ref{tab:lags}), indicating that human--model agreement did not depend substantially on the choice of lag. These lag adjustments were used only for human–model behavioral comparisons; inter-model correlations were calculated at zero lag.

\vspace{-.17in}

\subsection{Subject-Level fMRI Encoding}
\vspace{-.07in}
For each movie, behavioral measure (ToM or valence), and participant, we fit separate ridge-regression encoding models using baseline (no-context) ratings from each source (Human, LLaDA-V, Qwen3-VL, and Gemma 3) to predict BOLD responses across 114 ROIs. Each rating time series was shifted by a fixed 4-s hemodynamic delay.

All models used the same nested temporal cross-validation framework. To account for temporal autocorrelation in the movie and fMRI responses, each delay-shifted time series was divided into contiguous 16-s segments (20 TRs for TP and DM; 8 TRs for PC), yielding 11, 36, and 20 full segments for TP, DM, and PC, respectively. Outer folds were approximately leave-one-segment-out: each fold held out one segment, except that for TP and DM the 14 TRs (11.2 s) remaining after segmentation were held out together with one of the full segments, so that every TR was tested exactly once. Segments were always assigned intact to either the training or the test set; no additional temporal gap was inserted between training and test segments. Within each outer training set, five-fold inner cross-validation over the same segments selected $\lambda$ from a grid of 40 logarithmically spaced values between $10^{-3}$ and $10^{10}$. Multi-output ridge regression jointly predicted all ROI time series, with ROI-specific coefficients and a shared $\lambda$ selected by mean inner-cross-validation $R^2$ across ROIs. Input ratings were standardized using training-fold statistics, whereas fMRI responses were standardized independently within the training and test folds.

ROI-wise performance was measured by the Pearson correlation between predicted and held-out BOLD responses, emphasizing temporal alignment over scale and baseline differences. For each rating source (Human, LLaDA-V, Qwen3-VL, and Gemma 3), movie, and measure, group-level significance in each ROI was assessed with a one-sided sign-flip permutation test on subject-level correlations (5{,}000 permutations; $H_1$: mean $r > 0$). Benjamini--Hochberg FDR correction ($q < 0.05$) was applied across the 114 ROIs separately for each rating source, movie, and measure.

\vspace{-.17in}

\section{Results}
%  you may want to add some details on the three models' details and inference process
\vspace{-.03in}
\subsection{Machine ratings track human ratings of ToM and Valence}
\vspace{-.07in}

\begin{figure}[ht]
% \vspace{-.2in}
\includegraphics[width=\textwidth]{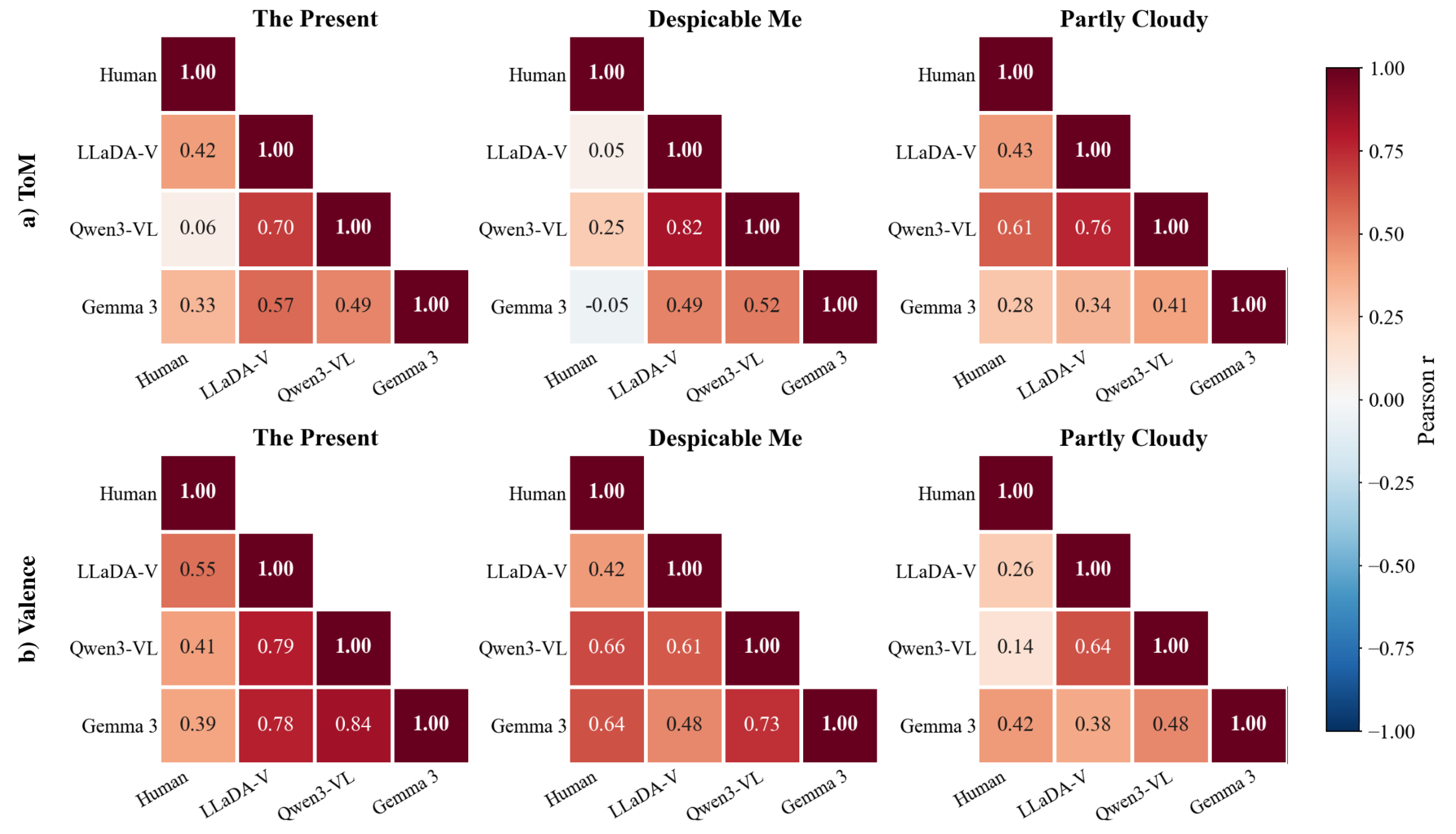}
\caption{Human–model and inter-model agreement in ToM (a) and valence (b) ratings across three animated movies. Each cell shows the Pearson correlation coefficient between the row and column judges (Human, LLaDA-V, Qwen3-VL, Gemma 3). Human--model values are lag-optimized correlations ($r^*$, exploratory; see Table~\ref{tab:lags}); inter-model values are unshifted. Diagonal values reflect self-correlation. Colors are centered at $r=0$, with dark red/dark blue denoting positive/negative correlations.}
\vspace{-.2in}
\label{fig1}
\end{figure}

% Across all 18 human-model matrix cells (3 movies $\times$ 3 models $\times$ 2 measures) in 
Fig.~\ref{fig1} shows Pearson correlations between human and MLLM ratings, as well as inter-model agreement. Overall, the MLLMs showed higher agreement with one another than with human ratings. Human--model agreement was generally higher for valence than for ToM: averaged across all three models and movies, mean $r$ was $0.434$ for valence versus $0.265$ for ToM.
This asymmetry varied across models. Gemma 3 showed the largest difference, with the highest mean human agreement for valence ($r=0.485$) but the lowest for ToM ($r=0.186$), including a near-zero, slightly negative correlation on DM ($r=-0.05$). LLaDA-V and Qwen3-VL showed a milder version of the same pattern, with higher agreement with humans for valence than for ToM. These results suggest that MLLM-generated ratings more closely capture human affective judgments than human mentalizing judgments, while inter-model agreement indicates greater consistency among models than with human ratings.

\begin{figure}[ht]
% \vspace{-.17in}
\centering
\includegraphics[width=\textwidth]{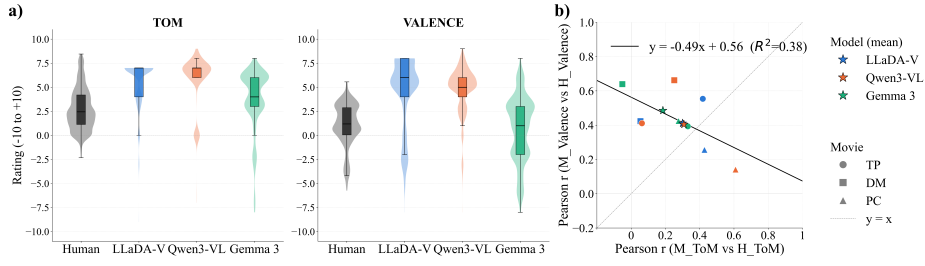}
\caption{\textbf{Rating distributions and human–model agreement.}
\textbf{(a)} Distributions of continuous, frame-wise ToM (left) and valence (right) ratings across three movies for human raters and three MLLMs. 
\textbf{(b)} Human–model Pearson correlations ($r^{*}$) for ToM ($x$-axis) and valence ($y$-axis). Colors denote models, shapes denote movies, and stars indicate model means. The solid line shows the least-squares fit.} 
\vspace{-.25in}
\label{fig2}
\end{figure}

In Fig.~\ref{fig2}(a), the three MLLMs differed substantially in their use of the rating scale. Human ratings were concentrated within a relatively narrow, central portion of the $-10$ to $+10$ range, whereas LLaDA-V and Qwen3-VL tended to assign higher scores than human raters, particularly for ToM. Gemma 3’s rating distributions were closer to those of humans, although differences remained.

% In Fig.~\ref{fig2}(a), the three models differed substantially in how they used the available rating scale. Human ratings were concentrated within a relatively narrow and central portion of the $-10$ to $+10$ range, whereas LLaDA-V and Qwen3-VL produced broader distributions extending more frequently toward the extremes for both ToM and Valence. In contrast, Gemma 3 showed a distributional spread more comparable to that of the human raters and did not exhibit the same systematic tendency toward extreme scores.

% Despite these differences in calibration, the three MLLMs were substantially more similar to one another than to human raters. Across the 18 human--model comparisons, the mean best-lag Pearson correlation was $r=0.349$, whereas the corresponding mean across the 18 model--model comparisons was $r=0.609$. Thus, the models shared considerably more temporal structure with one another than with human judgments, suggesting the presence of common model-specific biases or representational tendencies that are at least partly distinct from those underlying human ratings.

Across MLLMs, alignment with human valence judgments was also generally stronger than alignment with human ToM judgments (Fig.~\ref{fig2}(b)). Stars indicate mean correlations across the three movies.
Seven of the nine model--movie combinations fell above the identity line, indicating stronger alignment with human valence than ToM judgments. A least-squares fit suggested a negative trend between ToM and valence alignment ($y = -0.49x + 0.56$, $R^2 = 0.38$), although this trend should be interpreted cautiously given only nine non-independent observations.
\vspace{-.17in}

\subsection{Does temporal context improve model--human alignment?}
\vspace{-.08in}

\begin{figure}[ht]
\includegraphics[width=\textwidth]{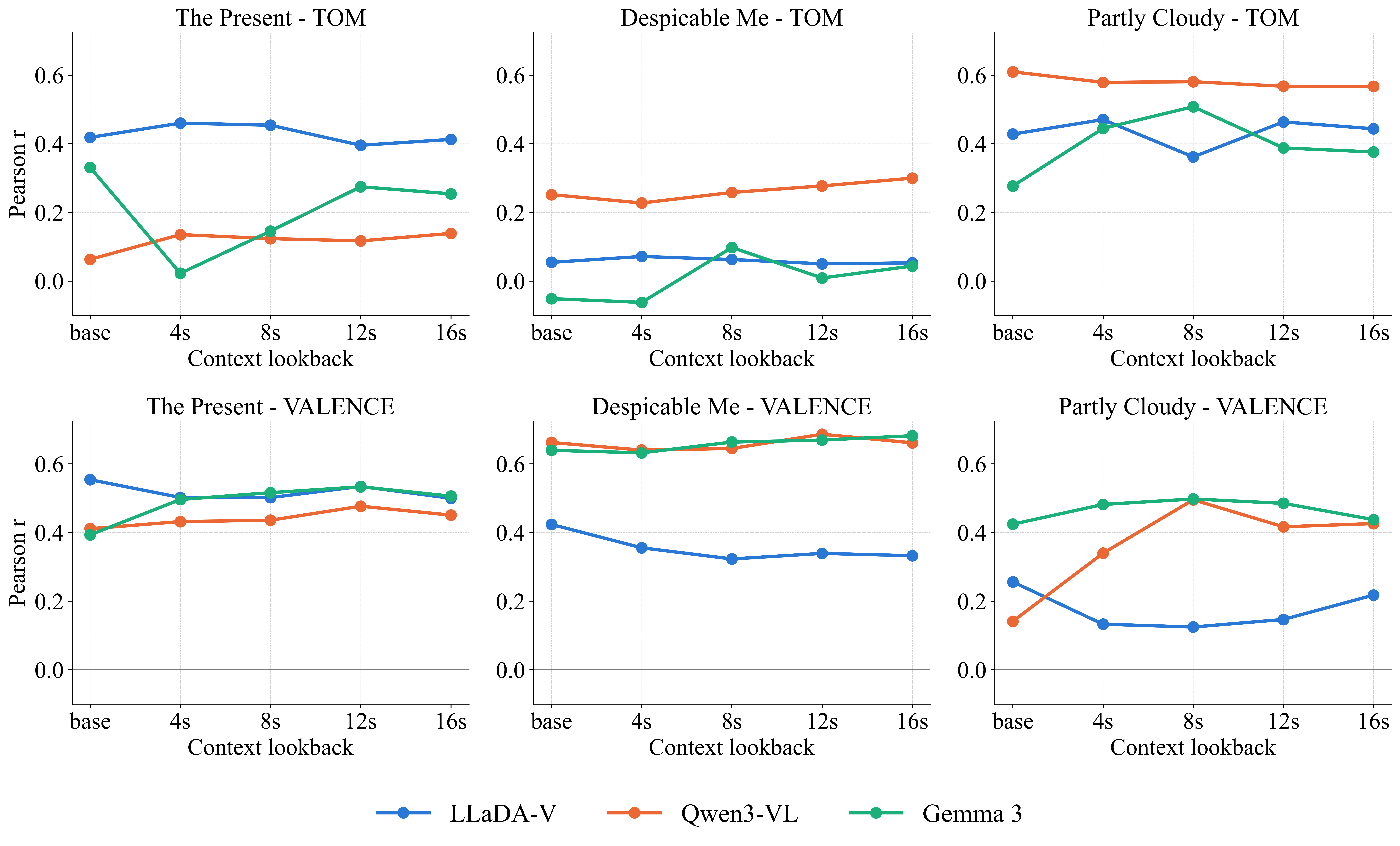}
\vspace{-.29in}
\caption{Effect of temporal context on human-model alignment. Pearson $r^{*}$ between human and model ratings as a function of the length of narrative context (previous scene descriptions) injected into the rating prompt, from no context (``base'') to 16 seconds of lookback, shown separately for each movie and measure.} 
\vspace{-.2in}
\label{fig3}
\end{figure}

In Fig.~\ref{fig3}, Qwen3-VL and Gemma 3 generally showed improved or comparable alignment with human ratings when preceding scene descriptions were added, particularly at intermediate lookback windows of approximately 8--12~s. Context improved human alignment more consistently for valence and benefited both ToM and valence in PC, while limited gains in some conditions may reflect already high correlations in the no-context condition. In contrast, LLaDA-V showed little improvement across context conditions.
% \vspace{-.2in}

% One possible explanation is that these descriptions provided limited information beyond the current frame, particularly when earlier events were redundant or less relevant to the ongoing scene. Alternatively, LLaDA-V may have incorporated contextual information without weighting or interpreting it in a way that better matched human judgments. Stable human--model correlations therefore do not necessarily indicate that context was ignored. Because longer windows contained the same number of descriptions sampled over a wider interval, they also did not guarantee more useful narrative information. These findings suggest that the benefit of context depends on both its relevance and how it influences the model's judgments; the present design cannot isolate the contribution of generative architecture.

An exception was Gemma 3’s marked decrease in ToM alignment under the 4-s context condition on TP. Inspection of early frames (approximately 8–11 s) revealed stronger mental-state attributions despite low human ToM ratings. With preceding context, Gemma 3 referred to earlier frames when interpreting the current scene (“combined with the previous moments… allowing me to understand”), whereas the no-context model expressed uncertainty (“lack of context… creates a barrier to understanding”). This contrast suggests that preceding descriptions may sometimes encourage over-interpretation of ambiguous scenes, potentially contributing to reduced alignment with human judgments.

% Second, LLaDA-V already showed relatively weak human alignment in several no-context conditions; for example, its correlation on \emph{Despicable Me}/ToM was approximately $r=0.05$--$0.10$, compared with approximately $r=0.25$--$0.30$ for Qwen3-VL and Gemma 3. This lower baseline may limit the extent to which additional context can improve alignment. 
% Importantly, these observations are consistent with an architectural contribution but do not establish it causally, because model architecture, scale, training data, and baseline performance were not independently controlled. A stronger test would require comparing diffusion-based and autoregressive models matched more closely in model capacity and baseline accuracy.

\vspace{-.15in}

\subsection{Do human and machine ratings predict brain activity?}
\vspace{-.1in}

\begin{figure}[ht]
% \vspace{-.07in}
\includegraphics[width=\textwidth]{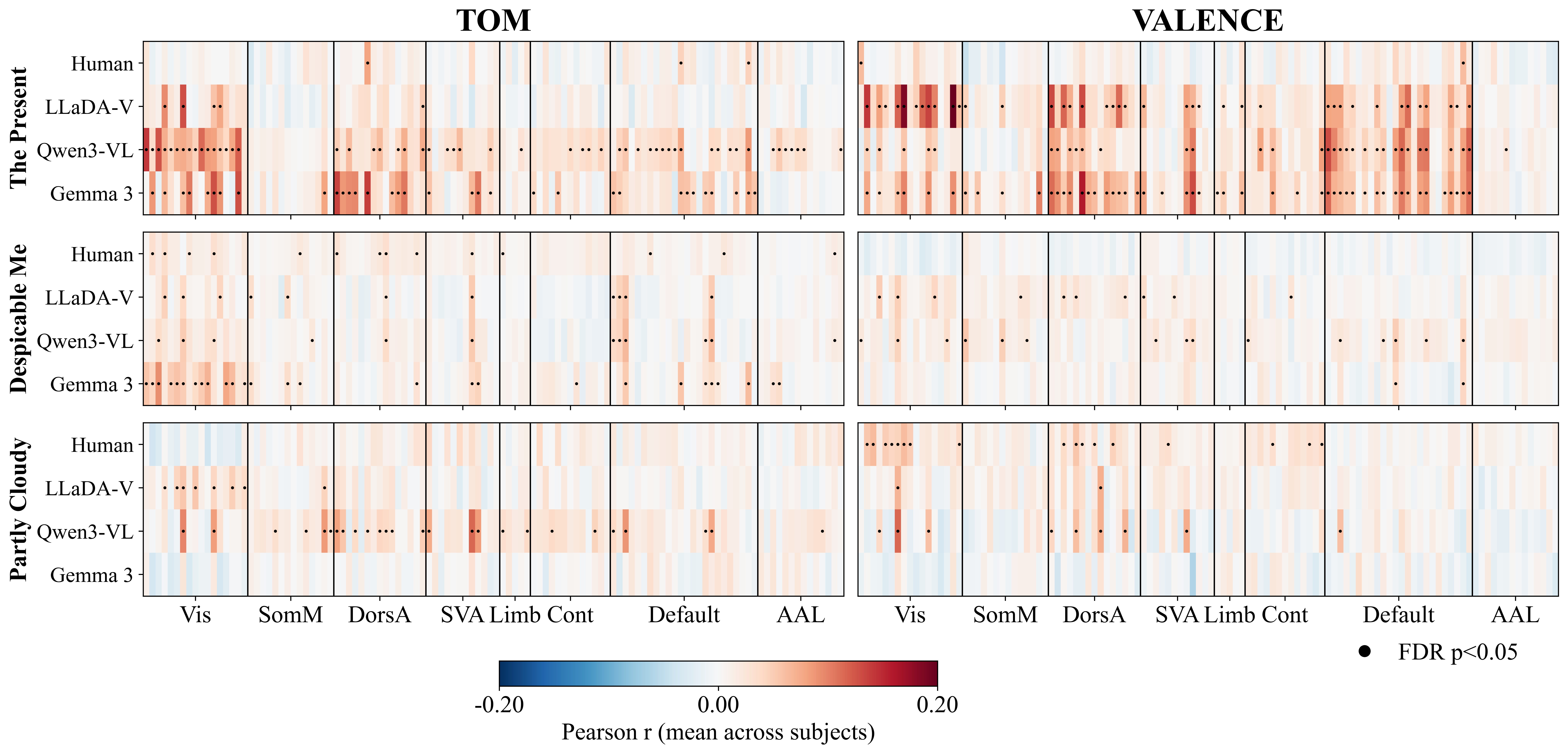}

\caption{Mean Pearson $r$ between predicted and held-out fMRI responses per ROI for human ratings and three models. Rows show three movies; columns show ToM and valence. ROIs are grouped by canonical cortical networks and AAL subcortical regions. Black dots indicate FDR-significant ROIs ($q<0.05$).
}
\label{fig4}
\vspace{-.24in}
\end{figure}

% \vspace{-.12in}
The heatmaps showed pronounced variation in encoding performance across movies, measures, and rating sources (Fig.~\ref{fig4}). Overall, TP exhibited stronger positive correlations and more widespread FDR-significant encoding, particularly for MLLM-generated ratings. Significant regions were concentrated in the visual, dorsal-attention, and default-mode networks, with additional involvement of the salience/ventral-attention network. Visual and dorsal-attention regions were prominent for both ToM and valence, whereas default-mode involvement was especially extensive for valence. Within TP, Qwen3-VL and Gemma 3 showed widespread significant ToM encoding, while all three MLLMs showed broad significant valence encoding. In TP, more ROIs reached significance for MLLM ratings than for human ratings, although this pattern did not generalize to every movie and measure; for PC valence, more ROIs reached significance for human ratings than for any individual MLLM. Because each source was tested against zero separately, these counts describe where encoding was reliably above chance for each source and do not indicate that one rating source predicted brain responses better than another.

\begin{figure}[t]
\vspace{-.2in}
\includegraphics[width=.98\textwidth]{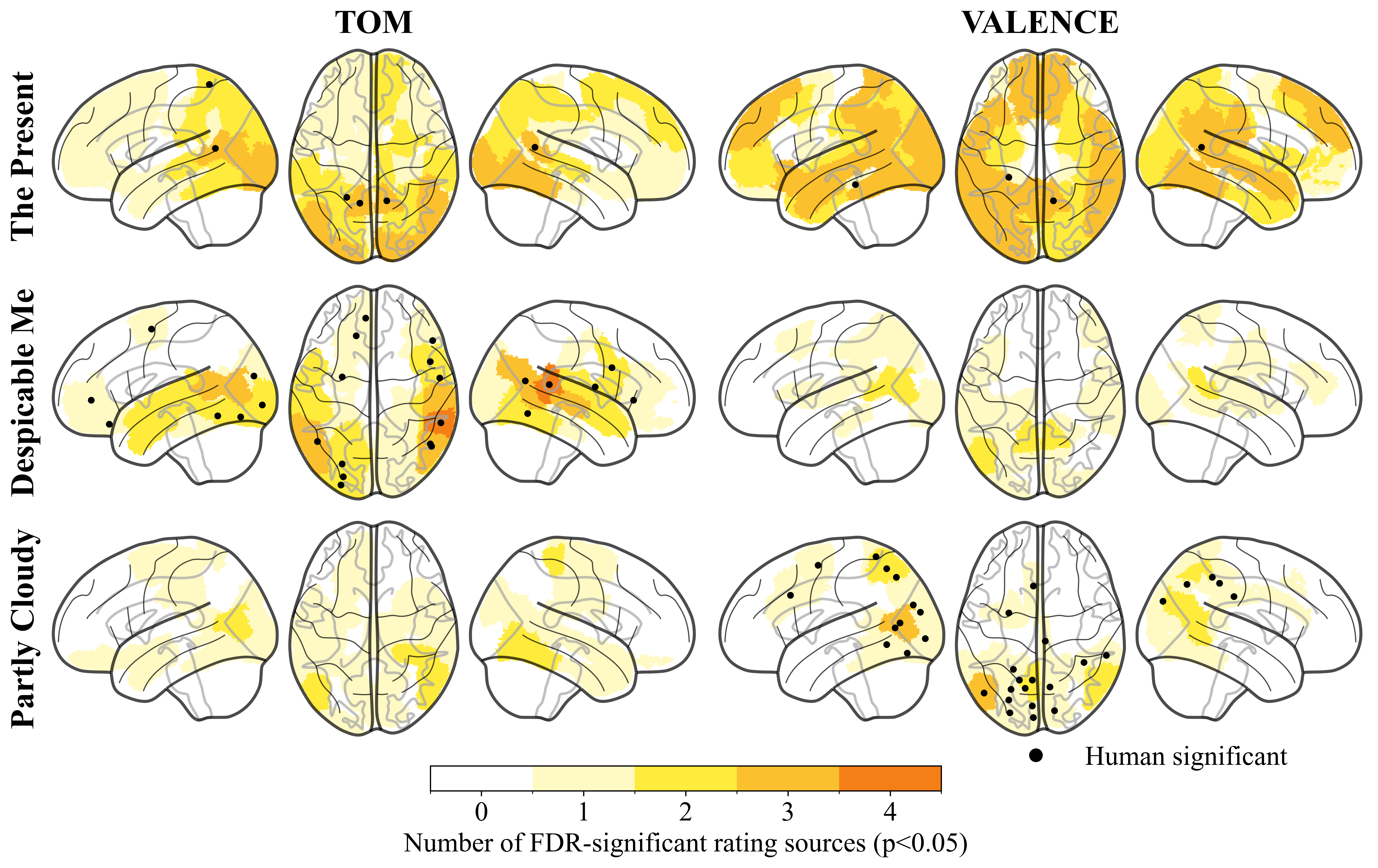}
\caption{\textbf{Spatial overlap of FDR-significant encoding across rating sources.}
Glass-brain projections were generated using Nilearn~\cite{abraham2014machine}, showing the number of rating sources (Human, LLaDA-V, Qwen3-VL, and Gemma~3) with significant encoding performance (FDR $q<0.05$) in each of the 100 Schaefer cortical ROIs. White indicates no significant sources; progressively darker yellow-to-orange shades indicate 1--4 significant sources (14 AAL subcortical ROIs were excluded). Black dots mark the centroids of ROIs significant for human ratings. ROIs significant for at least one MLLM but not for human ratings reflect model-only encoding; counts of two or more indicate overlap across sources.
}
\label{fig5}
\vspace{-.24in}
\end{figure}

The spatial overlap maps revealed the most extensive convergence across rating sources in TP (Fig.~\ref{fig5}). Because overlap counts do not indicate which sources are involved, we distinguish overlap that includes human ratings from overlap exclusively among MLLMs. In TP, most overlap occurred among MLLMs. For ToM, two-source overlap involved visual and dorsal-attention regions, together with portions of the default-mode and salience/ventral-attention networks, and was almost entirely between Qwen3-VL and Gemma 3. Three-source overlap was concentrated in visual regions, where it involved all three MLLMs, and in bilateral posterior cingulate/precuneus, where human ratings overlapped with Qwen3-VL and Gemma 3. For valence, three-source overlap extended broadly across the default-mode network and additionally involved visual, dorsal-attention, and salience/ventral-attention regions; it was confined to the three MLLMs except in the right posterior cingulate/precuneus, where human ratings were one of the three significant sources. No four-source overlap was observed in TP.

Overlap was more spatially restricted in the other movies, but human ratings contributed to a larger share of it. In DM, ToM showed two- and three-source overlap across several networks, much of which included human ratings (visual, somatomotor, and dorsal-attention regions), together with a focal region of four-source overlap in the right temporo-occipito-parietal salience/ventral-attention network, the only ROI significant for human ratings and all three MLLMs. Valence showed only limited two-source overlap in visual and default-mode regions, exclusively among MLLMs, as human valence ratings yielded no significant ROIs. In PC, ToM overlap was limited to two sources (LLaDA-V and Qwen3-VL) in visual and somatomotor regions, again without human ratings. For valence, two-source overlap occurred in two dorsal-attention regions, one involving human ratings and Qwen3-VL and another involving LLaDA-V and Gemma 3, and a localized three-source overlap in visual cortex included human ratings, LLaDA-V, and Qwen3-VL.

\vspace{-.12in}

\section{Discussion}
\vspace{-.03in}
We evaluated MLLMs as annotators of dynamic social and emotional content against human ToM and valence ratings and movie-evoked brain responses. MLLM ratings captured aspects of temporal variation in human judgments despite systematic distributional differences. LLaDA-V and Qwen3-VL tended to assign higher ratings than humans, whereas Gemma 3 produced ratings closer to human levels. Models nevertheless agreed more with one another than with humans, indicating limited correspondence. Human-model alignment also varied across measures, with stronger alignment for valence than ToM. The stronger valence alignment is consistent with evidence that multimodal models approximate human emotional judgments of images \cite{romeo2025artificial}. Visible expressions and scene composition may provide accessible emotional cues, whereas mental-state judgments often require information beyond the current frame. Nevertheless, emotion reasoning can also require contextual inference, and language models show uneven abilities in cognitive appraisal and emotional understanding \cite{yeo2025beyond,lv2026emotional}. Multimodal benchmarks similarly reveal difficulties integrating narrative evidence for ToM reasoning \cite{villa2025moments}. Because models saw only visual frames, missing dialogue may also matter: on dialogue-rich DM, ToM alignment was weakest while valence alignment remained high. Together, these factors may explain the asymmetry.

Providing preceding scene descriptions improved alignment in some conditions, but benefits varied across models and movies, potentially depending on both the relevance of the information and how models used it. The diffusion-based LLaDA-V showed little improvement, whereas the autoregressive Qwen3-VL and Gemma 3 benefited more consistently from preceding context.
Earlier descriptions may clarify unfolding interactions, whereas redundant, irrelevant, or inaccurate information may offer little benefit or weaken alignment. Models may also incorporate context without weighting or interpreting it in ways that better match human judgments. Thus, LLaDA-V's relatively stable human-model correlations do not necessarily indicate that context was ignored. Because description counts were fixed, longer lookback windows tested the temporal reach of sampled context rather than its quantity and did not necessarily provide more useful narrative information. The present design cannot isolate the contribution of generative architecture to these differences.

Machine ratings also predicted held-out BOLD responses, complementing evidence that AI-derived semantic annotations contain brain-predictive information \cite{li2026mapping}. Significant encoding was particularly widespread in TP, prominently involving visual, dorsal-attention, and default-mode networks. However, extensive overlap primarily reflected convergence among models; overlap involving human ratings was more restricted.

Several limitations remain. First, predictive associations establish neither human-like social understanding nor specificity to ToM or valence, as both measures may co-vary with broader factors such as character presence, visual motion, or arousal. Second, individual frames and scene descriptions may also miss contextual and auditory cues relevant to ToM and valence judgments, and future work should test whether richer audiovisual inputs improve behavioral and neural alignment. Third, human ratings came from young adults, whereas the imaging participants were largely children and adolescents; adult ratings may not reflect how younger viewers perceive mental states or emotional tone, and should be viewed as an adult reference. Finally, optimal human--model lags were similar for ToM and valence and largely consistent across models within each movie, clustering around 4~s for DM and PC but shorter for TP ($\leq 1.6$~s); lags were less stable for DM ToM. Although lag optimization changed correlations only modestly relative to unshifted values, the optimized estimates remain exploratory and should be validated on held-out data.

Overall, MLLMs provide scalable, brain-predictive annotations of social and emotional content, with human alignment varying across models, measures, movies, and temporal context; establishing what these annotations capture beyond correlated perceptual features remains an open question.

\vspace{-.15in}

\section{Acknowledgment}
\vspace{-.12in}

We acknowledge partial funding support from DOE SEA-CROGS project (DE-SC0023191) and AFOSR project (FA955024-1-0231) to Y.H., Z.Z., and M.X.; the U.S. National Institutes of Health (R15MH125332) and the New Jersey Governor’s Council for Medical Research and Treatment of Autism (CAUT25BRP005) to X.D. We thank George B. Hanna and Rounak Rahman for assistance with behavioral data collection, and NJIT High Performance Computing (HPC) facility.
\vspace{-.3in}

\bibliographystyle{splncs04}
\bibliography{ref}  




% LNCS Homepage, \url{http://www.springer.com/lncs}, last accessed 2023/10/25
% \end{thebibliography}
\end{document}